# Online Adaptive Platoon Control for Connected and Automated Vehicles via Physics Enhanced Residual Learning

Peng Zhang, Heye Huang*, Hang Zhou, Haotian Shi, Keke Long, Xiaopeng Li

Department of Civil and Environmental Engineering, University of Wisconsin-Madison, Madison, WI, 53706, USA

*Correspondence
E-mail address: pzhang257@wisc.edu (P. Zhang), hhuang468@wisc.edu (H. Huang), hzhou364@wisc.edu (H. Zhou), hshi84@wisc.edu (H. Shi), klong23@wisc.edu (K. Long), xli2485@wisc.edu (X. Li)

**Abstract**
This paper introduces a physics enhanced residual learning (PERL) framework for connected and automated vehicle (CAV) platoon control, addressing the dynamics and unpredictability inherent to platoon systems. The framework first develops a physics-based controller to model vehicle dynamics, using driving speed as input to optimize safety and efficiency. Then the residual controller, based on neural network (NN) learning, enriches the prior knowledge of the physical model and corrects residuals caused by vehicle dynamics. By integrating the physical model with data-driven online learning, the PERL framework retains the interpretability and transparency of physics-based models and enhances the adaptability and precision of data-driven learning, achieving significant improvements in computational efficiency and control accuracy in dynamic scenarios. Simulation and robot car platform tests demonstrate that PERL significantly outperforms pure physical and learning models, reducing average cumulative absolute position and speed errors by up to 58.5% and 40.1% (physical model) and 58.4% and 47.7% (NN model). The reduced-scale robot car platform tests further validate the adaptive PERL framework's superior accuracy and rapid convergence under dynamic disturbances, reducing position and speed cumulative errors by 72.73% and 99.05% (physical model) and 64.71% and 72.58% (NN model). PERL enhances platoon control performance through online parameter updates when external disturbances are detected. Results demonstrate the advanced framework's exceptional accuracy and rapid convergence capabilities, proving its effectiveness in maintaining platoon stability under diverse conditions.

**Keywords**：Physics enhanced residual learning, Connected and automated vehicles, Centralized platoon control, Online adaptive control

# 1. Introduction

Connected and automated vehicle (CAV) platoon represents a significant advancement in intelligent transportation systems through advanced cooperative control algorithms, offering prospects for enhancing road capacity and improving traffic safety (Z. Huang et al., 2023; Zhou et al., 2021). The platoon can maintain safe inter-vehicle distances and coordinating braking or acceleration to improve overall traffic flow. In real traffic environments, the platoon system faces unpredictable conditions such as extreme weather, dynamic scenarios, and curved roads, resulting in a complex and non-linear system (Huang et al., 2020; Kennedy et al., 2023). The increasing complexity and non-linearity demands robust, adaptable control mechanisms for platoons.

Existing CAV platoon control methods can be categorized into three types (Li et al., 2022, 2018): physical model-based control, such as rule-driven and optimization methods; learning-based approaches, such as multi-agent collaborative control based on reinforcement learning and deep learning; and hybrid approaches, which integrate learning methods on the foundation of physical modeling.

**Physical model-based control.** Physical model-based control strategies for CAV platoons predominantly rely on classical control theories, leveraging well-established physical rules of vehicle dynamics and behavior (Caruntu et al., 2016). Methods like adaptive cruise control (ACC) and cooperative adaptive cruise control (CACC) are key, with ACC using sensor data for individual vehicle control and CACC building on this through V2X communications for coordinated group control (Li et al., 2016; Rubió-Massegú et al., 2013). However, while these methods provide a solid foundation for safety and reliability, they often overlook the complex, dynamic interactions between vehicles in dense urban settings. Linear and nonlinear control theories, including sliding mode control (SMC) (Mohd Zaihidee et al., 2019) and model predictive control (MPC) (Caruntu et al., 2016), offer more flexibility by considering future states and multiple objectives like safety and fuel efficiency. In typical longitudinal platoon system, studies have developed distributed nonlinear MPC suitable for vehicle platoons with a unidirectional topology (Wang and Su, 2022). These controllers perform merging-in and merging-out operations by tracking desired speeds and maintaining safe inter-vehicle distances, thereby achieving collision avoidance. Du et al. (Zhang and Du, 2023) developed a control strategy that enables a mixed flow platoon to efficiently navigate signalized intersections, demonstrating the sequential and switching feasibility, as well as the Input-to-State stability, of the hybrid MPC system. Mao et al.(Mao et al., 2023) developed a distributed tube model predictive control method that guarantees the string stability of heterogeneous vehicle platoons under external disturbances and controller saturation. This proposed control strategy significantly enhances disturbance rejection performance and substantially reduces the online computation burden(Hu et al., 2023). Despite their robust theoretical bases and highly interpretability, these methods often oversimplify the complex dynamics of real-world driving, especially in heterogeneous platoon systems. This can lead to significant control errors, causing the system's output to deviate notably from ideal control and prediction scenarios (Kennedy et al., 2023; Kianfar et al., 2015).

**Learning-based control.** Learning-based control methods enable online adaptations to dynamic environments through data-driven approaches (Huang et al., 2024; Shi et al., 2023b). Machine learning and reinforcement learning have emerged as powerful tools for developing control policies directly from traffic data, effectively capturing characteristic nonlinear relationships and evolving dynamics in urban traffic (Liu et al., 2020; Wang et al., 2024). These methods excel in environments where the vehicle dynamics and interactions are too complex for traditional models to handle accurately. For instance, deep reinforcement learning (DRL) and neural networks (NN) have been utilized to optimize platoon strategies in real-time, adapting to changes in traffic flow and signal timings without relying on predefined models (Gao et al., 2024; Wang et al., 2023). Shi et al. (Shi et al., 2023a) proposed a cooperative control strategy for CAVs in mixed traffic environments, utilizing DRL to segment the platoon into multiple subsystems for centralized management. Li et al. developed an adaptive platoon-based intersection control model, named INTEL-PLT, which utilizes learning-based control to optimize multiple dynamic objectives (Li et al., 2023). Additionally, the concept of multi-agent reinforcement learning (MARL), examined by Busoniu et al.(Busoniu et al., 2006), has gained widespread acceptance for controlling platoons of networked CAVs. Xu et al. presented a MARL-based control method for CAVs within a mixed platoon, thereby enhancing the model's capability to manage spatial relationships among platoon members and improving platooning control effectiveness (Xu et al., 2024). Multi-agent systems further extend these capabilities by enabling decentralized decision-making within CAV platoons, leveraging NNs to learn dynamic models from vast datasets, thereby enabling scalable and flexible traffic management (Luo et al., 2021; Yan et al., 2022). Learning-based control methods can learn complex controls can be achieved through continuous exploration of the environment. While they often lack interpretability and transparency, which poses challenges in understanding the control processes and dynamic mechanisms involved in multi-vehicle autonomous driving. Moreover, their dependence on substantial amounts of high-quality training data also presents challenges, especially in safety-critical applications and rare driving scenarios where data is scarce.

**Hybrid control.** Recognizing the limitations of purely physical or learning-based methods, recent research has shifted towards hybrid approaches that combine the strengths (Bahavarnia et al., 2024; Jiang and Keyvan-Ekbatani, 2023). Physics-informed machine learning (PIML) techniques, such as physics-informed neural network (PINN) exemplify this trend, where traditional control models inform the feature engineering and training processes of machine learning algorithms (Jang et al., 2023; Latrach et al., 2024). This integration enhances the learning efficiency and generalizability of the models under limited data conditions typical in novel traffic scenarios (O'Connell et al., 2022). For instance, we aim to leverage a physical model as the foundation to achieve approximately 90% of the desired control performance from scratch, and then employ neural networks and other learning-based methods to capture and refine the remaining 10% induced by dynamic factors. These hybrid methods are designed to maintain the interpretability and reliability of physical models while enhancing adaptive performance provided by machine learning in complex real-world scenarios (H. Huang et al., 2023). However, the application in CAV platoon system is still very limited. Directly integrating PIML can lead to insufficient exploration of the physical model, rendering it difficult to accurately develop a fundamental physical model that reflects system complexity, thereby introducing bias. Furthermore, PINN is often trained from the ground up rather than learning residuals on top of an existing foundation model. Consequently, bolstering the robustness of residual dynamic learning requires additional training data, and when data are scarce, its effectiveness diminishes.

To address this gap, this paper introduces an online physics enhanced residual learning (PERL) controller for CAV platoons. It uses a physics-based controller to model vehicle dynamics and employs a NN-based residual controller correcting dynamic residuals. **Fig. 1** compares the PERL model with three existing models: the fixed physics model, the NN pure learning model, and the PINN model. The fixed physics model, reliant on offline control, incurs increased positional errors. The NN model improves control through data but lacks interpretability and is heavily data-dependent. The PINN model, merging physical and data-driven features, offers adaptability but underperforms with limited data. Our proposed PERL framework innovatively combines physical models with data-driven NN methods. It utilizes prior knowledge from the physical model and employs residual learning to correct physical model predictions, enhancing control precision in scenarios with platoon dynamics disturbances. Our contributions are:

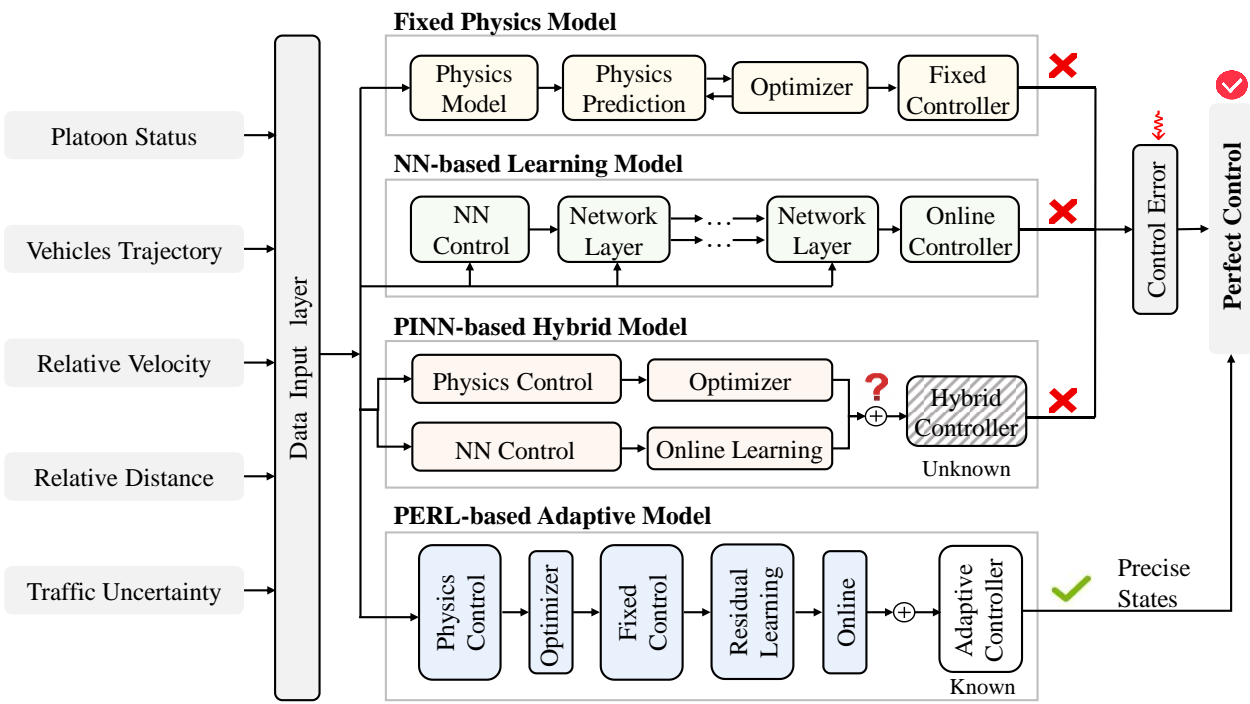

**Figure 1.** Comparison of different model structures. (a) Fixed physics model, (b) NN-based learning model, (c) PINN-based hybrid model and (d) PERL-based adaptive model.

(1) We introduce a novel PERL framework that seamlessly integrates a physical model of vehicle dynamics with a neural network-based residual learning approach. This integration allows for the correction of residuals induced by complex vehicle dynamics, enhancing the adaptability and precision of CAV platoon control without sacrificing the interpretability inherent in physics-based models.

(2) By employing tests on a scaled-down robot car platform, we have validated the superiority of adaptive models in platoon control performance, particularly regarding error metrics and loss values. These tests not only highlight the superiority of PERL over fixed physical models and purely learning-based models but also underscore its feasibility and superiority in practical applications.

(3) The PERL framework enhances the capability to adjust to external disturbances through online parameter updates, ensuring stability in platoon control. Experimental results showcase the framework's exceptional accuracy and rapid convergence capabilities in maintaining platoon stability under diverse conditions, proving its effectiveness in dynamic environments.

The remainder of this paper is arranged as follows. In Section 2, we propose the model framework to illustrate the proposed PERL. In Section 3, our proposed PERL control method from the aspect of physical controller, and the residual NN learning component is introduced. Then, we describe design typical scenarios for simulation verification and quantitative experiment results are shown in Section 4. Section 5 tests the proposed PERL through the reduced-scale platform. Conclusions are given in Section 6.

## 2. Framework

As shown in **Fig. 2**, the proposed vehicle platoon control system consists of three main modules: vehicle platoon disturbance, inherent physical model, and residual learning model. The controller's input consists of uncertain vehicle platoon states with multiple disturbances, while the output includes control actions with residual compensation for all vehicles. (1) Vehicle platoon disturbance module: this module accounts for disturbances such as non-linearity, dynamic coupling, vehicle interactions, and external friction. these are captured and used in the control model to understand and adjust for deviations in platoon behavior. (2) Inherent physical model module: This module generates the reference trajectory by considering vehicle dynamics, reference states, and physical control signals. The optimizer uses vehicle dynamics to generate predictions, with the vehicle's physical parameters and control inputs jointly determining its motion. Serving as the core of the controller, this module ensures that each vehicle adheres to the intended path based on its dynamic and physical characteristics. (3) Residual learning model module: this module compensates for residual errors not addressed by the physical model. A NN is trained with a small amount of dynamic data to learn and adjust control signals in real time, compensating for discrepancies between the ideal and actual vehicle behaviors, thereby enhancing accuracy and system robustness. Through this adaptive learning process, the controller allows for appropriate driving speeds and real-time control adjustments, minimizing deviations from the reference trajectory while maximizing stability. Together, these modules enable the controller to effectively handle dynamic changes and external disturbances, ensuring precise and stable control for all vehicles in the platoon.

Specifically, PERL fundamentally differs from PINN in principle, as illustrated in **Fig. 1** (b) and (c). The physical model is not merely a guide for the learning process but a core component of the PERL output. By incorporating prior knowledge into PERL, the residual learning model focuses on predicting the residuals of the physical model as corrections. PERL adopts a more intuitive and rational principle, possessing distinctive advantages: (1) lower bias risk due to the physical model; (2) efficient learning by reducing dimensions and concentrating on residuals; (3) lower data requirements; (4) improved interpretability and transparency—crucial qualities for cyber-physical systems like autonomous vehicles. The advantages enable the improvement of platoon control in uncertain traffic conditions.

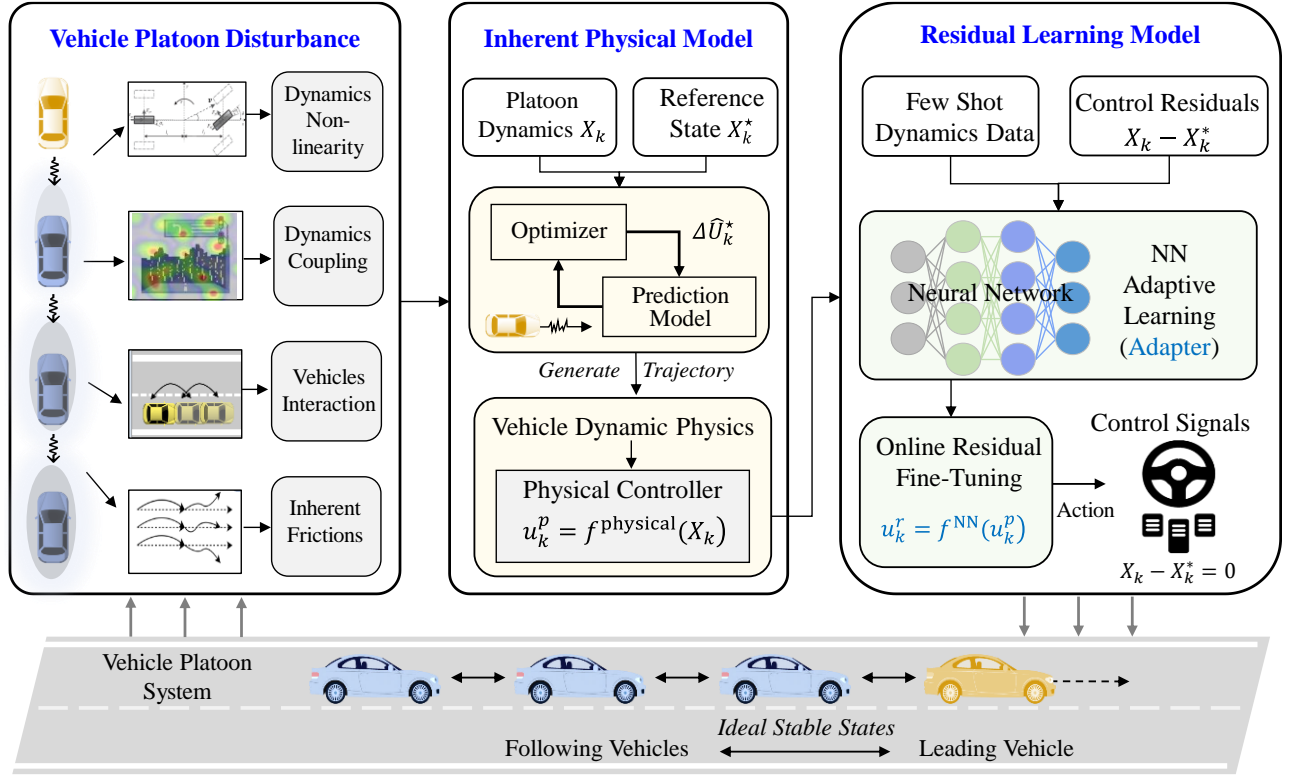

**Figure 2.** The integrated CAVs platoon control framework.

## 3. Adaptive platoon control

The PERL controller framework streamlines into two core components: (1) The foundational physical-based controller that accounts for platoon dynamics in centralized control. (2) A supplementary learning module that embeds a residual feedback mechanism into the physical model, employing NN for adaptive online correction of model inaccuracies and disturbances. This dual system facilitates precise speed regulation and dynamic control responses, ensuring vehicles adhere closely to their trajectories for optimal platoon stability.

*3.1 Vehicle platoon system dynamics*

The platoon, consisting of $I + 1$ homogeneous vehicles including a leader and $I$ followers, is a crucial component of the control system. The main task of the platoon control is to use V2V communication to form a flexible mixed platoon from individual vehicles on the road, achieving vehicle coordination through mutual cooperation. According to the hierarchical structure, platoon control is divided into centralized control and distributed control. Centralized control relies on a central controller to collect and process data from all vehicles, generating control commands to coordinate the movement of each vehicle. This approach can optimize and coordinate the entire system. In this paper, we employ the centralized control for real vehicle platoon system.

According to the vehicle longitudinal dynamics, the linear model for a single vehicle $i \in \mathcal{I} = \{1,2,\dots,I\}$ at time step $k \in \mathbb{Z}$ with constant sampling interval $\Delta t$:

$$\begin{cases} p^i_{k+1} = p^i_k + v^i_k \Delta t + \dfrac{1}{2} a^i_k \Delta t^2 \\ v^i_{k+1} = v^i_k + a^i_k \Delta t \\ a^i_{k+1} = -\dfrac{\Delta t}{\tau^i} v^i_k + \dfrac{\Delta t}{\tau^i} s^i_k = -\dfrac{\Delta t}{\tau^i} v^i_k + \dfrac{\alpha^i \Delta t}{\tau^i} u^i_k + \dfrac{\beta^i \Delta t}{\tau^i} \end{cases} \quad (1)$$

where $p^i_k$ denotes the position, $v^i_k$ denotes the speed, $a^i_k$ denotes the acceleration.

Then, we can get the state space model for vehicle $i$ at time $k$:

$$x^i_{k+1} = A^i x^i_k + B^i u^i_k + C^i \quad (2)$$

where $x^i_k = \left[p^i_k, v^i_k, a^i_k\right]^\top$ is the state information for vehicle $i$,

$$A^i = \begin{bmatrix} 1 & \Delta t & \dfrac{\Delta t^2}{2} \\ 0 & 1 & \Delta t \\ 0 & -\dfrac{\Delta t}{\tau^i} & 0 \end{bmatrix}, B^i = \begin{bmatrix} 0 \\ 0 \\ \dfrac{\alpha^i \Delta t}{\tau^i} \end{bmatrix}, C^i = \begin{bmatrix} 0 \\ 0 \\ \dfrac{\beta^i \Delta t}{\tau^i} \end{bmatrix} \quad (3)$$

We assume all vehicles in the platoon are homogeneous ($A^i = A, B^i = B, \tau^i = \tau$). We then define:

$$X_k = [p^1_k, \ldots, p^I_k, v^1_k, \ldots, v^I_k, a^1_k, \ldots, a^I_k]^\top \quad (4)$$

$$U_k = [u^1_k, \ldots, u^I_k]^\top \quad (5)$$

such that the platoon dynamics are:

$$X_{k+1} = A_I X_k + B_I U_k \quad (6)$$

where $A_I = A \otimes E_I, B_I = B \otimes E_I, \otimes$ is the Kronecker operator, and $E_I$ is the $I$ dimensional elementary matrix.

Define the change in control actions $\Delta U_k$ from the previous control action $U_{k-1}$:

$$\Delta U_k = U_k - U_{k-1} = [\Delta u^1_k, \ldots, \Delta u^I_k]^\top \quad (7)$$

where $\Delta u^i_k$ is the change in control action for vehicle $i$.

To predict the platoon's state for the next $N$ steps, we introduce the predicted state value of the platoon at time $k+n$ for $n \in \mathcal{N} = \{1, \ldots, N\}$ from the measured state value at time $k$, denoted as $\hat{X}_{k+n|k}$, with the prediction window defined as:

$$\mathcal{X}_k = \left[\hat{X}^\top_{k+1|k}, \ldots, \hat{X}^\top_{k+N|k}\right]^\top \quad (8)$$

For the predicted value of the platoon controller as:

$$\Delta \widehat{U}_k = \left[\Delta \widehat{U}^\top_{k|k}, \ldots, \Delta \widehat{U}^\top_{k+N-1|k}\right]^\top \quad (9)$$

where from the measurement at time $k$ the predicted applied control at time $k+n$ is:

$$\widehat{U}_{k+n|k} = \widehat{U}_{k+n-1|k} + \Delta \widehat{U}_{k+n|k} \quad (10)$$

$$\Delta \widehat{U}_{k+n|k} = \left[\Delta \hat{u}^1_{k+n|k}, \ldots, \Delta \hat{u}^I_{k+n|k}\right]^\top \quad (11)$$

The state prediction of the platoon $\mathcal{X}_k$ can be written as a linear combination of the current state $X_k$, the previously applied control $U_{k-1}$ and the predicted change in control $\Delta \widehat{U}_k$:

$$\mathcal{X}_k = \Phi X_k + \lambda U_{k-1} + \Gamma \Delta \widehat{U}_k \quad (12)$$

where

$$\Phi = \begin{bmatrix} A_I \\ \vdots \\ A_I^N \end{bmatrix}, \lambda = \begin{bmatrix} A_I^0 B_I \\ \vdots \\ (A_I^{N-1} + \cdots + A_I^0) B_I \end{bmatrix} \quad (13)$$

$$\Gamma = \begin{bmatrix} B_I & \cdots & 0 \\ \vdots & \ddots & \vdots \\ (A_I^{N-1} + \cdots + A_I^0) B_I & \cdots & B_I \end{bmatrix} \quad (14)$$

Denote the reference state as $p_k^{i\star}, v_k^{i\star}, a_k^{i\star}$, and

$$X_k^\star = [p_k^{1\star}, \dots, p_k^{I\star}, v_k^{1\star}, \dots, v_k^{I\star}, a_k^{1\star}, \dots, a_k^{I\star}]^\top \quad (15)$$

$$\mathcal{X}_k^\star = [(X_{k+1}^\star)^\top, \dots, (X_{k+N}^\star)^\top]^\top \quad (16)$$

Consider for each vehicle $i \in \mathcal{I}$, the absolute position, velocity, and acceleration errors as the difference between the current state and the reference state:

$$\begin{cases} \tilde{p}_k^i = p_k^i - p_k^{i*} \\ \tilde{v}_k^i = v_k^i - v_k^{i*} \\ \tilde{a}_k^i = a_k^i - a_k^{i*} \end{cases} \quad (17)$$

For the entire platoon, these errors can be written as:

$$X_k - X_k^* = [\tilde{p}_k^1, \dots, \tilde{p}_k^I, \tilde{v}_k^1, \dots, \tilde{v}_k^I, \tilde{a}_k^1, \dots, \tilde{a}_k^I]^\top \quad (18)$$

Then denote $\hat{p}_{k+n|k}^i, \hat{v}_{k+n|k}^i, \hat{a}_{k+n|k}^i$ as the prediction error where the subscript indicates the state prediction at time $k + n$ given the state at time $k$.

*3.2 Optimized MPC for platoon control*

Centralized control requires a central controller to optimize vehicle control at each timestep, facing challenges from intertwined vehicle states and controls, and the need to forecast complex, long-term traffic dynamics due to dimensionality and disturbances. MPC, a closed-loop method, predicts system behavior for a defined future period and optimizes control laws within constraints, applying only the first-step control. Repeated each timestep, MPC effectively adjusts to control errors and uncertainties, making it well-suited for CAV Platoon control. The formulation of the MPC controller, featuring a finite prediction horizon of $N$ steps, is as follows:

$$J(k, N) = \min \sum_{n=0}^{N-1} \left[ \sum_{i=1}^{I} q_1(\hat{p}_{k+n|k}^i)^2 + q_2(\hat{v}_{k+n|k}^i)^2 + q_3(\hat{a}_{k+n|k}^i)^2 + q_4(\Delta \hat{u}_{k+n|k}^i)^2 \right] \quad (19)$$

$$s.t.:$$
$$d_{\min} \leq p^{i-1} - p^i \leq d_{\max}, \quad \forall i \in \mathcal{I} \quad (20)$$
$$v_{\min} \leq v^i \leq v_{\max}, \quad \forall i \in \mathcal{I} \quad (21)$$
$$a_{\min} \leq a^i \leq a_{\max}, \quad \forall i \in \mathcal{I} \quad (22)$$

where $q_1, q_2, q_3, q_4$ correspond to errors in position, velocity, acceleration, and control inputs. $d_{\max}, d_{\min}, a_{\max}, a_{\min}, v_{\max}, v_{\min}$ are the maximum and minimum limits for space, acceleration, and velocity. Constraint (20) ensures platoon safety and distance limits; constraint (21) enforces road speed limits; constraint (22) sets acceleration limits based on vehicle engine and braking capacities.

The problem can be written in the form of a quadratic program:

$$J(X_k, \Delta \hat{U}_k) = \Delta \hat{U}_k^\top (\Psi + \Gamma^\top \Omega \Gamma) \Delta \hat{U}_k + 2(\Phi X_k + \lambda U_{k-1} - \mathcal{X}_k^\star)^\top \Omega \Gamma \Delta \hat{U}_k \tag{23}$$

s.t.:
$$\bar{G} \Gamma \Delta \hat{U}_k \leq -\bar{G}(\Phi X_k + \lambda U_{k-1}) - \bar{g} \tag{24}$$

where $\Omega = \text{diag}\{Q, \ldots, Q, 0\}, \Psi = \text{diag}\{R_\Delta, \ldots, R_\Delta\}$ are block diagonal matrices,

$$Q = \begin{bmatrix} q_1 E_I & 0 & 0 \\ 0 & q_2 E_I & 0 \\ 0 & 0 & q_3 E_I \end{bmatrix}, R_\Delta = q_4 E_I \tag{25}$$

$$\bar{G} = \text{diag}[\check{G}, \ldots, \check{G}], \bar{g}^\top = [g^\top, \ldots, g^\top], \tag{26}$$

$$\check{G} = \begin{bmatrix} \mathfrak{T}_I & 0 & 0 \\ -\mathfrak{T}_I & 0 & 0 \\ 0 & -E_I & 0 \\ 0 & E_I & 0 \\ 0 & 0 & -E_I \\ 0 & 0 & E_I \end{bmatrix}, g = \begin{bmatrix} 1_{I-1} d_{\min} \\ -1_{I-1} d_{\max} \\ 1_I v_{\min} \\ -1_I v_{\max} \\ 1_I a_{\min} \\ -1_I a_{\max} \end{bmatrix}, \tag{27}$$

where $\mathfrak{T}_I$ is a size $(I-1) * I$ Toeplitz matrix with $-1$ on the diagonal and $1$ on the first upper diagonal. Vectors $1_{I-1}$ and $1_I$ are columns of ones of size $(I-1)$ and $I$, respectively. Advanced solvers can rapidly solve this quadratic optimization problem. The optimal platoon control action minimizes the constrained finite horizon cost function.

$$\Delta \hat{U}_k^\star = \underset{\Delta \hat{U}_k}{\text{argmin}}\ J(X_k, \Delta \hat{U}_k) \tag{28}$$

where the first item $\Delta \hat{U}_{k|k}^\star$ will be the output control $u_k^p$ of the physical model.

### 3.3 The inherent vehicle physical model

For vehicle platoon system, to effectively model the nonlinear characteristics of platoon dynamics within processing constraints and without incurring substantial computational expenses, we utilize a model based on physical principles that captures the essence of these dynamics efficiently. The control command, expressed in terms of revolutions per minute (RPM), ensures the smooth and safe operation of the entire platoon by directly adjusting the duty cycle of the pulse width modulation (PWM), which ranges from 0% to 100%. This PWM duty cycle, in turn, linearly dictates the motor voltage, thereby determining the motor speed and allowing for precise regulation of the robotic system's movement, as shown in **Fig. 3**.

Ideally, the relationship between motor voltage and motor speed should be linear. However, experimental results indicate that this relationship is not entirely linear. The observed nonlinearity is primarily due to internal motor friction, transmission friction, voltage drops caused by motor resistance, and battery voltage drop. Consequently, the physical model can be divided into two parts: a linear part and an uncertain/nonlinear part that can be learned using PERL:

$$v_k^i = \alpha^i u_k^i + \beta^i + R \tag{29}$$

where $v_k^i$ is desired speed of vehicle $i$ at time step $k$, $u_k^i$ is the control input, $a$ is a constant known from the motor's characteristics, set at 615.4, and $\beta$ (motor rotation resistance) is a constant determined from experimental results. It is evident that adding a constant to the model improves its accuracy.

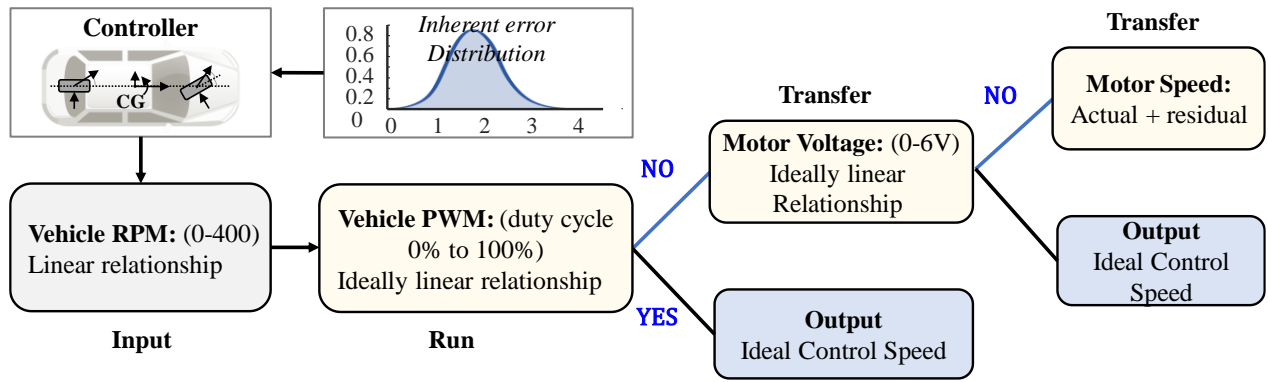

**Figure 3.** The control command for vehicle platoon system.

Based on the Root Mean Square Error (RMSE) calculated between the desired and actual speeds, the value of β should be set to 25. Generally, the residual between the physical model and desired speed is most noticeable in the low-speed and high-speed ranges, which is represented by R. This residual can be learned and corrected using PERL, improving the accuracy of the physical model. $R$ contains friction of transmission/ Back Electromotive Force/ Motor resistance, which will be hard to measure. Thus PERL could learn this part and then feed it back to physical model.

*3.4 Online adaptation based on residual learning*

Residual learning refines control outputs by addressing the "gap" between predicted and actual system states. Simplified motion models often introduce residual dynamic errors between actual outputs and intended commands. This process focuses on discrepancies between the desired speed from the physical model and the actual vehicle speed, guiding adjustments to the Direct Control Variable (DCV), which varies by platform—throttle and brake for full-sized vehicles or motor RPM for scale models. This study utilizes NN-based residual learning, with states and actions defined based on physical model outputs and their rate of change. To manage these continuous variables, the action space is discretized within the vehicle's acceleration range.

**Algorithm 1** illustrates the PELR flow adapted for vehicle platoon environment, which includes the physical model control steps and online adaptation based on residual learning. The Physical controller is enhanced for a centralized CAV platoon, utilizing vehicle speed as the control input and focusing on multi-objective collaborative optimization. The learning-based residual controller augments the MPC's prior knowledge and rectifies residuals caused by traffic disturbances. We will subsequently validate the algorithm's effectiveness through simulation experiments and a scaled-down car platform.

| | |
|---|---|
| **Algorithm 1:** Physical Enhanced Residual Learning for CAVs platoon control | |

**Input:** Ego vehicle $i \in \mathcal{I} = \{1,2,...,I\}$, initial vehicle platoon states $X_k = [p_k^1,...,p_k^I, v_k^1,...,v_k^I, a_k^1,...,a_k^I]^\top$, inertial delay $\tau^i$, vehicle reference state $[p_k^{i\star}, v_k^{i\star}, a_k^{i\star}]$, learning rate $\eta$, discount factor $\rho$, exploration rate $\epsilon$.

**Output:** Optimal vehicle platoon control $u_k^{i\,*}$

1: **for** $k \leftarrow 0$ *to* $n$ **do**   // Update real time steps
2: $\quad \mathcal{X}_k \leftarrow [\hat{X}_{k+1|k}^\top, ..., \hat{X}_{k+N|k}^\top]^\top$   // Initialize the predicted platoon state value
3: $\quad \Delta \widehat{U}_k \leftarrow [\Delta \widehat{U}_{k|k}^\top, ..., \Delta \widehat{U}_{k+N-1|k}^\top]^\top$   // Estimate the predicted value of the platoon controller
4: $\quad$ **for** $i \leftarrow 1$ *to* $I$ **do**   // Platoon length
5: $\quad\quad X_k - X_k^* \leftarrow [\tilde{p}_k^1,...,\tilde{p}_k^I, \tilde{v}_k^1,...,\tilde{v}_k^I, \tilde{a}_k^1,...,\tilde{a}_k^I]^\top$   // Calculate platoon errors from references
6: $\quad\quad$ Define $\hat{p}_{k+n|k}^i, \hat{v}_{k+n|k}^i, \hat{a}_{k+n|k}^i$   // Initialize the platoon prediction error
7: $\quad\quad$ **for** $n \leftarrow 0$ *to* $N$ **do**   // Update the prediction horizon
8: $\quad\quad\quad J(k,N) \leftarrow \min \sum_{n=0}^{N-1} \left[ \sum_{i=1}^{I} q_1^2 + q_2^2 + q_3^2 + q_4^2 \right]$   // Minimize cost
9: $\quad\quad\quad \Delta \widehat{U}_k^\star \leftarrow \underset{\Delta \widehat{U}_k}{\mathrm{argmin}}\ J(X_k, \Delta \widehat{U}_k)$   // Run MPC
10: $\quad\quad\quad s_k^i = \alpha^i u_k^i + \beta^i + R$   // Run the inherent vehicle physical model
11: $\quad\quad\quad$ **if** $X_k - X_k^*\ != 0$ **then**
12: $\quad\quad\quad\quad R \leftarrow residual\_model$   // Calculate the residual by residual_model
13: $\quad\quad\quad\quad R.\text{AddLayer}\ (64,\ \text{input\_dim}=3,\ \text{activation}=\text{'relu'})$   // Input layer: desired and actual speed, residual
14: $\quad\quad\quad\quad R.\text{Compile}\ (\text{optimizer}=\text{'adam'},\ \text{loss}=\text{'mse'})$ // Compile the model
15: $\quad\quad\quad\quad$ Train (X, y, epochs=100, validation_split=0.2) // Train the residual_model
16: $\quad\quad\quad\quad R\_pred \leftarrow residual\_model.\text{Predict}\ (X)$ // Predict residuals
17: $\quad\quad\quad\quad$ data ['Corrected_Speed'] $\leftarrow$ data ['Rob_1_Speed'] + $R\_pred$ () // Calculate the final predicted speed by correcting the actual speed
18: $\quad\quad\quad\quad$ Fast online adaptation
19: $\quad\quad\quad\quad$ End the algorithm, output the control strategy
20: $\quad\quad\quad\quad$ **break**
21: $\quad\quad\quad$ **else**
22: $\quad\quad\quad\quad u_k^{i\,*} \leftarrow u_k^i$   // Obtain the optimal strategy
23: $\quad\quad\quad$ **end if**
24: **return**: the optimal vehicle platoon control $u_k^{i\,*}$ with minimal control error

**Fig. 4** illustrates the dynamic residuals between the reference and predicted trajectory control in a CAV platoon system. It outlines the key phases of trajectory prediction and control, divided into past and future periods. The red line represents the reference trajectory, while the blue line depicts the predicted trajectory. Based on the outputs of the baseline physical model, residuals inevitably arise, which are mitigated through residual compensation using neural network (NN) learning. The PERL framework achieves 90% of the control effect via the baseline model and further incorporates NN-driven residual learning and compensation to reduce the remaining 10% of residual errors and disturbance impacts.

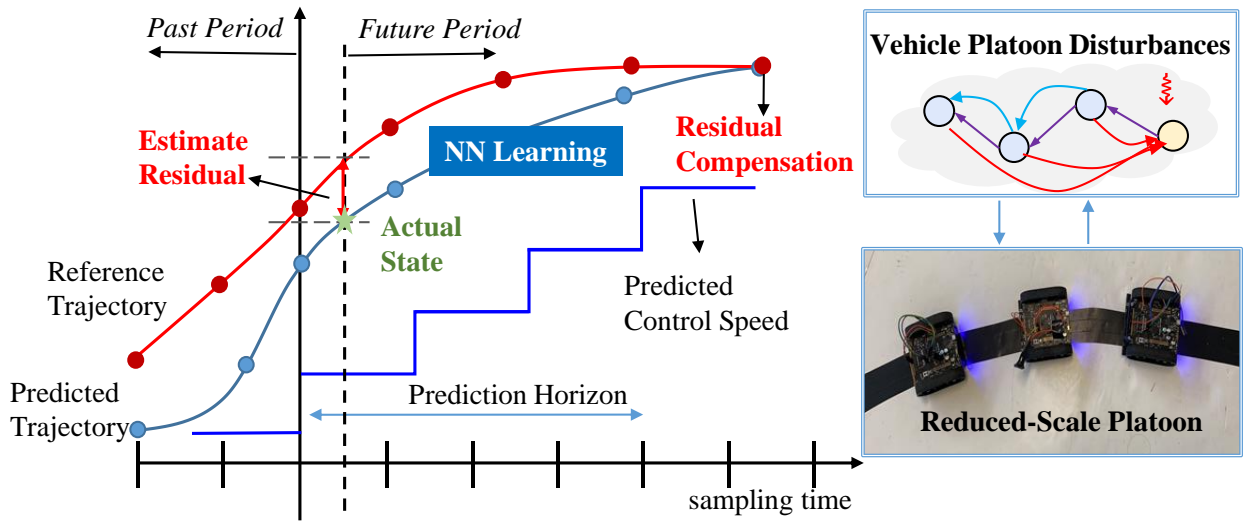

**Figure 4.**   The dynamic residuals between reference and predicted trajertory control in CAV platoon.

## 4. Simulation experiment and results

In this section, we assess the performance of the proposed PERL controller model within a simulation environment by conducting a comparative analysis against two baseline models. The simulation platform utilized is Python 3.8.18. The quadratic programming problem is solved using the Sequential Least Squares Programming (SLSQP) method, implemented through the "scipy.optimize" package in Python.

*4.1 Experiment setting and baselines*
In this simulation, we conFig.d a platoon of 5 vehicles for 30 seconds, with a $\Delta t = 0.1$ seconds. To test the effectiveness of the PERL controller under different scenarios, we designed two types of test trajectories. (1) Real-world platooning trajectories, sourced from the OpenACC dataset (Makridis et al., 2021). These trajectories are invaluable as they provide empirical data reflecting actual vehicle behavior in platooning scenarios, thereby offering a realistic benchmark for algorithmic performance. (2) Synthetic trajectories from car-following models. Specifically, the intelligent driver model (IDM) (Treiber et al., 2000) is applied to obtain the longitudinal trajectories. The models are particularly versatile due to its capacity to adjust headway parameters, thus allowing for the simulation of various traffic densities and driving behaviors. The formulation of the IDM is shown as follows:

$$a_k = a_{max} \left[ 1 - \left(\frac{v_k}{v_0}\right)^\delta - \left(\frac{s^*(v_k, \Delta v_k)}{s_k}\right)^2 \right] \quad (30)$$

where the definitions and values of the parameters are shown in **Table 1** and the desired minimum gap $s^*(v_k, \Delta v_k)$ is calculated as:

$$s^*(v_k, \Delta v_k) = s_0 + s_1 \sqrt{\frac{v_k}{v_0}} + Tv_k + \frac{v_k \Delta v_k}{2\sqrt{ab}} \quad (31)$$

For the first type of trajectories, we randomly extracted one trajectory as Scenario 1. For the second type of trajectory, another one trajectory is generated as Scenario 2. The Speed-Time Diagram of the two scenarios is shown in **Fig. 5**. By employing both types of trajectories, the study aims to cover a comprehensive range of driving conditions, enhancing the robustness and applicability of the developed algorithms.

**Table 1** Model parameters of the IDM used in the reference trajectory generation.

| Parameters | Typical value |
|---|---|
| Desired speed $v_0$ | 33.3 m/s |
| Safety time headway $T$ | 1.6s |
| Maximum acceleration $a_{max}$ | 0.73 m/s² |
| Desired deceleration $b$ | 1.67 m/s² |
| Acceleration exponent $\delta$ | 4 |
| Jam distance $s_0$ | 2 m |
| Jam distance $s_1$ | 0 m |

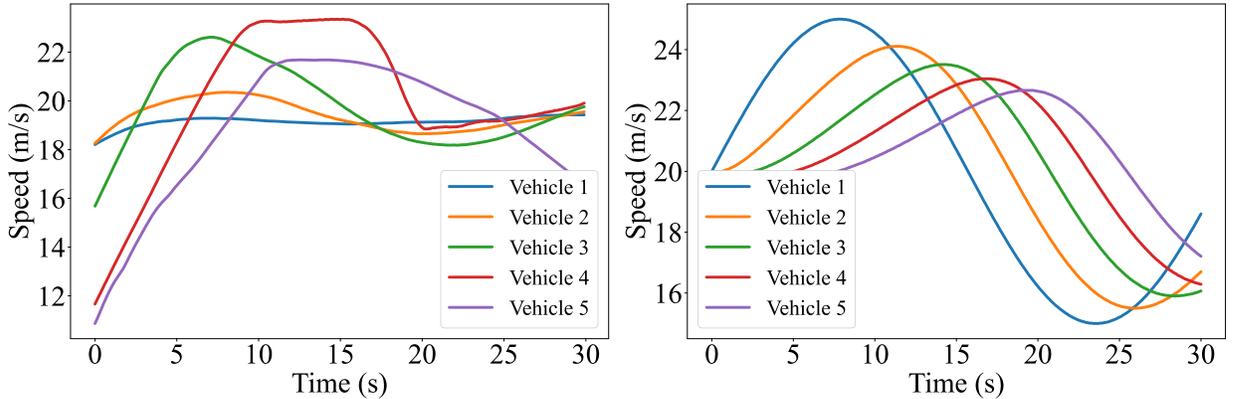

**Figure 5.** Example of the OpenACC and IDM trajectories.

The parameters in the MPC constraints are set as follows: $d_{min} = 5\text{m}, d_{max} = 80\text{m}, a_{max} = 5\text{m/s}^2, a_{min} = -5\text{m/s}^2, v_{max} = 50\text{m/s}, v_{min} = 5\text{m/s}$. To account for control output errors during RPM and velocity transfer, two error models were used: affine error $u_k^a = 1.1u_k - 3 +$ $x$, $x \sim N(0,1)$, and quadratic error $u_k^q = 0.01u_k^2 + u_k - 3 + x$, $x \sim N(0,1)$. The online learning updated the NN every 20 time steps (2 seconds) using experimental data. To evaluate the effectiveness of the PERL method, the results are compared with the MPC with physical model and the PERL controller. These methods were tested across two scenarios and two error types, totaling four evaluations.

### 4.2 Simulation results and analysis

Table 2 details the cumulative and maximum absolute velocity errors for the MPC with physical model and the online PERL method across four tests. We use four metrics to measure the performance of the two methods: the cumulative absolute error and the maximum absolute error for position and velocity, denoted as $CAE_p$, $MAE_p$, $CAE_v$, and $MAE_v$, respectively. The 'Gap' represents the difference in errors when compared to those obtained by the PERL method. The results demonstrate that the PERL method consistently outperforms the MPC with the physical model and the MPC with NN across all four tests. Specifically, the average cumulative absolute position errors for the PERL method are 58.5% lower compared to the MPC with the physical model and 58.4% lower compared to the MPC with NN. Furthermore, the average gap in cumulative absolute velocity errors between the PERL method and the other two methods is 40.1% and 47.7%, respectively, indicating that the PERL method significantly outperforms the alternatives. In terms of maximum absolute error, the error gaps for position between the PERL method and the other methods are substantial, at 53.3% and 57.7%, respectively. However, the gap in maximum absolute velocity error is relatively smaller, averaging 2.1% and 17.4%, respectively. Notably, in Scenario 2 with quadratic error, the maximum absolute error produced by the PERL method is even greater than that of the other two methods. The reasons behind this phenomenon will be further analyzed in the subsequent paragraph.

**Table 2** Simulation results for the three control models in four tests.

| Test | Scenario 1 (affine error) | | | Scenario 1 (quadratic error) | | |
|---|---|---|---|---|---|---|
| Method | Physics | NN | PERL | Physics | NN | PERL |
| $CAE_p$ | 3689 | 2409 | <u>1884</u> | 5503 | 9194 | <u>2892</u> |
| $GapCAE_p$ | 48.9 | 21.8 | <u>0.0</u> | 47.4 | 68.5 | <u>0.0</u> |
| $CAE_v$ | 268.6 | 200.5 | <u>154.0</u> | 367.6 | 631.3 | <u>273.8</u> |
| $GapCAE_v$ | 42.6 | 23.2 | <u>0.0</u> | 25.5 | 56.6 | <u>0.0</u> |
| $MAE_p$ | 7.0 | 5.0 | <u>3.7</u> | 8.1 | 16.2 | <u>5.4</u> |
| $GapMAE_p$ | 47.2 | 26.4 | <u>0.0</u> | 33.4 | 66.9 | <u>0.0</u> |
| $MAE_v$ | 0.6 | 0.5 | <u>0.5</u> | 0.8 | 1.1 | <u>0.6</u> |
| $GapMAE_v$ | 19.5 | 7.7 | <u>0.0</u> | 21.3 | 41.3 | <u>0.0</u> |
| Test | Scenario 2 (affine error) | | | Scenario 2 (quadratic error) | | |
| Method | Physics | NN | PERL | Physics | NN | PERL |
| $CAE_p$ | 12953 | 5089 | <u>1985</u> | 3752 | 10016 | <u>1769</u> |
| $GapCAE_p$ | 84.7 | 61.0 | <u>0.0</u> | 52.9 | 82.3 | <u>0.0</u> |
| $CAE_v$ | 838.5 | 345.1 | <u>155.1</u> | 310.6 | 629.0 | <u>277.0</u> |
| $GapCAE_v$ | 81.5 | 55.1 | 0.0 | 10.8 | 56.0 | 0.0 |
| $MAE_p$ | 21.4 | 9.0 | 3.7 | 6.9 | 16.3 | 3.4 |
| $GapMAE_p$ | 82.8 | 58.8 | 0.0 | 49.8 | 78.9 | 0.0 |
| $MAE_v$ | 1.1 | 1.3 | 0.6 | 0.9 | 1.3 | 1.7 |
| $GapMAE_v$ | 48.3 | 56.2 | 0.0 | -80.6 | -35.6 | 0.0 |

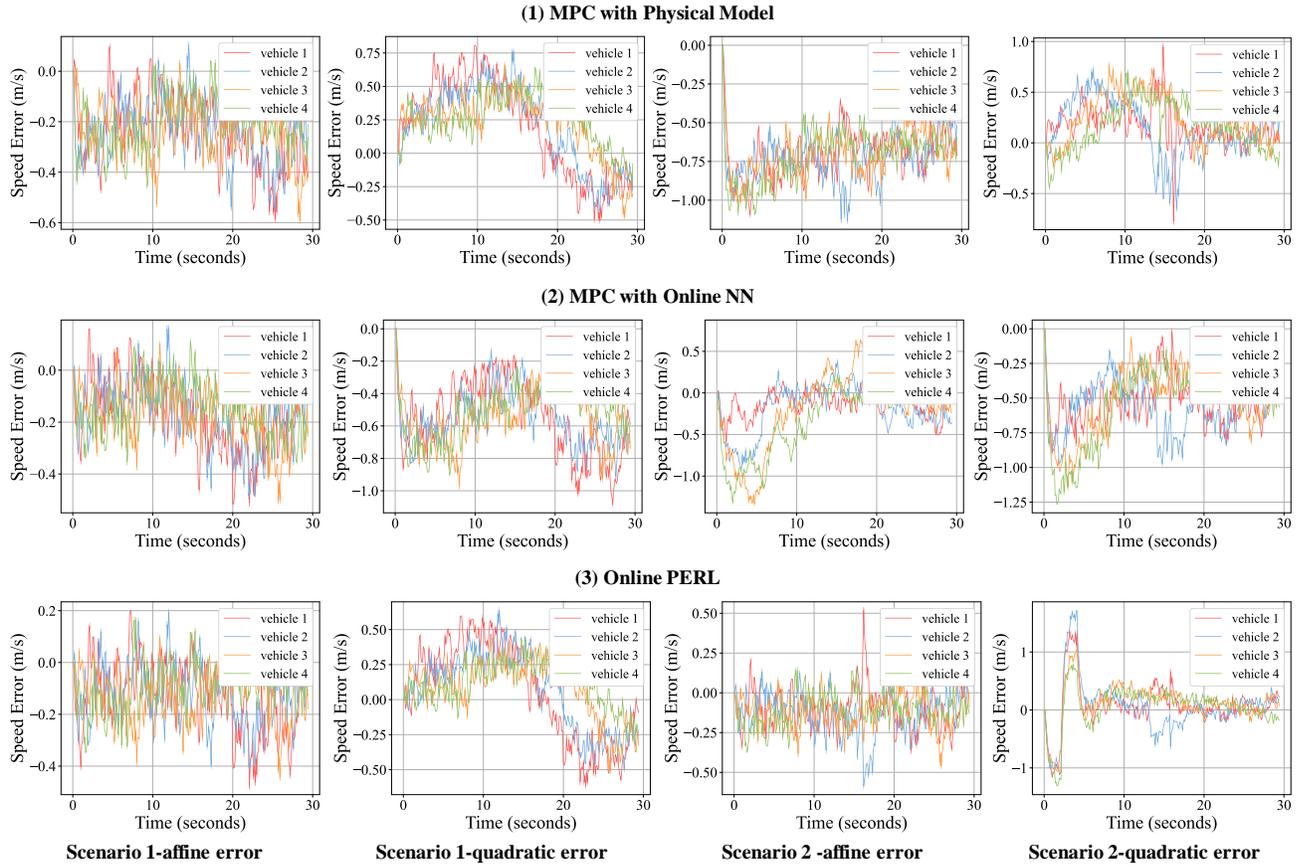

**Figure 6.** The comparison of experimental results the three control models.

The variations in velocity error across all trajectories are depicted in **Fig. 6**, providing an explanation for the results presented in **Table 2**. The results from both **Table 2** and **Fig. 6** confirm that the online PERL method significantly outperforms the MPC with the physical model and NN. Additionally, **Fig. 6** demonstrates that, in all four tests, the errors for PERL eventually stabilize near zero. This effect is particularly evident in Scenario 2 with quadratic error. For the online PERL controller, the initial lack of sufficient training data results in a large negative error. However, as training progresses, the residual learning component effectively learns and corrects the error, reducing the velocity error below zero and compensating for the initial position error caused by the negative velocity error at the outset. This phenomenon explains the maximum absolute error observed for the PERL method in **Table 2**. After training, the error stabilizes within the ±0.3 range, validating the effectiveness of online learning. In contrast, the MPC with NN approach, lacking the support of a physical model, continues to exhibit significant error even after training, indicating that achieving results comparable to those of PERL would require a larger volume of training data.

However, the current simulations, using simplified control environments and error forms, may not mirror real-world conditions accurately, thus necessitating further real-world testing to validate the PERL methods' performance, e.g. in the real vehicle platforms.

## 5. Reduced-scale platform test

For the experiment, we utilize a reduced-scale robot car platform shown in **Fig. 7** to test the effectiveness of the proposed PERL. Reduced-scale robot cars offer complete controllability, thereby mitigating safety risks during experimental trials. They serve as efficient proxies for validating theoretical models prior to their application in full-scale vehicles. The components of the reduced-scale platform include Pololu Zumo 32U4 robots, infrared sensors for distance measurement, a PC equipped with an Intel Core i5 processor, 8 GB RAM, and macOS for intensive computations, Wi-Fi chips to facilitate communication, and a circular test track. Through this reduced-scale platform, we can derive velocity and acceleration and test the effectiveness of PERL.

*5.1 Reduced-scale platform*

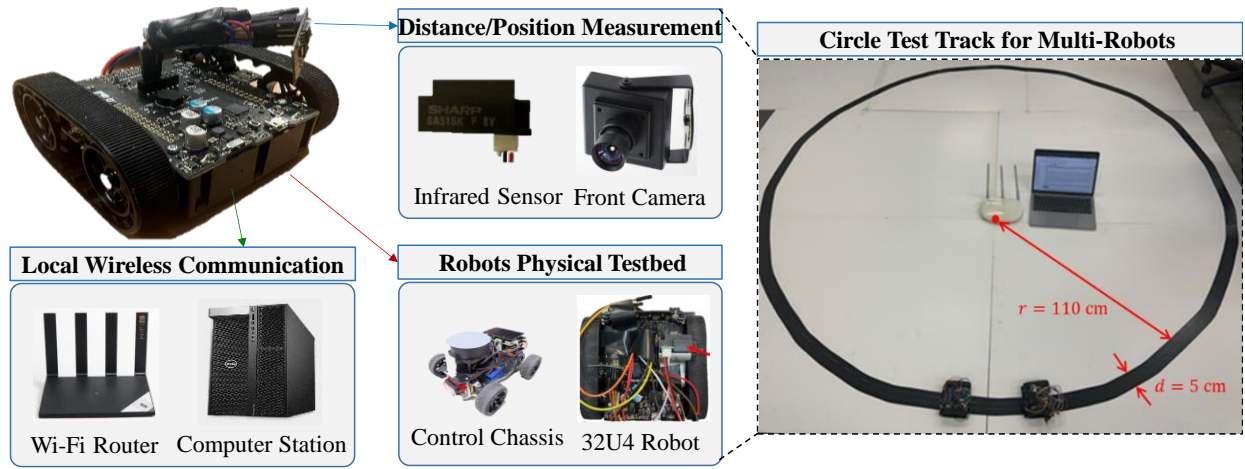

**Figure 7.**   Reduced-Scale Platform and the Test track.

As shown in **Fig. 7**，in this reduced-scale platform, we divide it into 4 parts, including: 1) Robots: The experimental setup featured a Pololu Zumo 32U4 robot, equipped with a 75:1 HP gear ratio and advanced line sensors for precise line tracking. Key components include an Arduino microcontroller for ease of programming, additional serial port for communication upgrades, and dual motors with encoders for accurate position and speed monitoring. 2) Distance/Position Measurement: An infrared sensor, GP2Y0A51SK0F, mounted on the robot provides rapid and accurate distance measurements by converting analog signals to digital. This facilitates effective collision avoidance controlled by a calibrated proportional-derivative (PD) controller. 3) Wireless Communication: The robot uses an ESP8266 Wi-Fi chip, chosen for its low power consumption and IP/TCP protocol support, enabling robust communication with the PC via a Wi-Fi router. This setup allows for the exchange of real-time speed data and vehicle-specific trajectories. 4) Test Track: The test track is a ring-shaped path with a 110 cm inner radius and 5 cm width, designed to optimize line tracking via sensor-calibrated reflectivity contrast between the track materials.

### 5.2 Experiment setting

Table 3 details the parameter settings for the online adaptation algorithm, where both physical model and online learning update parameters every 80 milliseconds. A count threshold (Cth) of 5 allows for parameter updates every 0.4 seconds, independent of road condition changes.

**Table 3** Parameters of online adaptation.

| Parameters | Optimal | Parameters | Optimal |
|---|---|---|---|
| Layer | 64 | Input dim | 3 |
| Epochs | 100 | Validation split | 0.2 |
| Verbose | 1 | Loss | MSE |
| Activation | relu | Optimizer | ADAM |

To assess the control accuracy of various algorithms, we deployed three Pololu Zumo robots, each equipped with consistent dynamic models, to form a multi-robot platoon that autonomously navigated a predetermined circular track (as shown in Fig. 8). During the experiment, the robots' states, such as position and speed, were continuously recorded. Fig. 8(a) illustrates the platoon system, where the robots, under dynamic disturbances, followed the circular trajectory, testing the platoon's ability to maintain stable control and precise tracking. Fig. 8(b) presents the X-Y position plots of the robot trajectories, which validate the platoon's performance under the PERL and baseline models (physical, NN) through multiple trajectory control trials conducted on the circle test track. Fig. 8(c) demonstrates the reduced-scale platform testing, which includes a series of evaluations designed to optimize the platoon's stability and trajectory control for the reduced-scale robot models, offering valuable insights into the scalability of the proposed control strategies for more complex systems.

The evaluations are divided into two stages to comprehensively assess the performance of the algorithms: 1) Preliminary testing, assessing stability control through pure trajectory tracking in both single and multi-robot platoons, focusing on responsiveness and coordination; 2) Introducing disturbances, where dynamic challenges are introduced to the circle scenario in order to test trajectory control under real-world conditions with external disruptions. Fig. 8(d) highlights various control errors, including disturbance errors, robot dynamics-related control errors, and position and speed control errors, to evaluate the overall performance and robustness of the algorithms. The errors offer insights into the system's ability to recover from perturbations and maintain the desired trajectory.

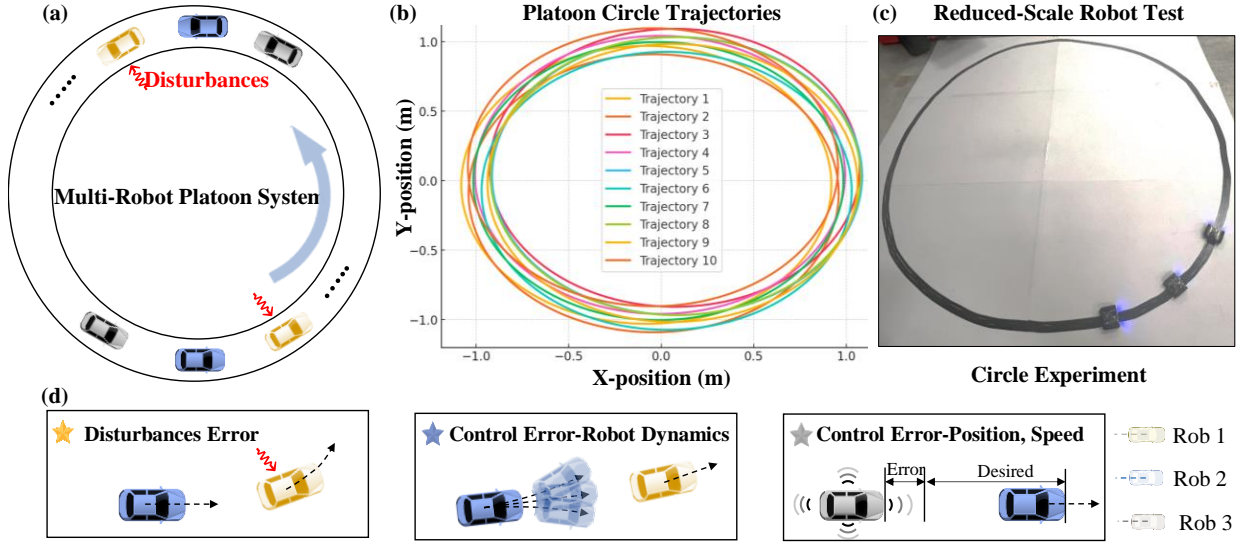

**Figure 8.** Robot platoon circle trajectories for 10 laps with a target velocity.

*5.3 Experiment results and analysis*

Fig. 9 presents the experimental results of speed and position control for a single robot under three models: the Fixed Physical Model, the Online Neural Network (NN) Model, and the Online PERL Framework. Each graph shows the robot's speed (desired speed as a dashed line, actual speed as a solid line) and position (desired position as a dashed line, actual position as a solid line) over time steps. The Fixed Physical Model exhibits significant fluctuations, especially between time steps 300 and 400, indicating slow response and imprecise adjustments. In contrast, both the Online NN and PERL models follow the desired speed more closely, with PERL offering smoother variations, highlighting its superior accuracy and responsiveness.

Regarding position control, the Fixed Physical Model keeps the robot's position near the desired trajectory, with slight deviations later on. The Online NN and PERL models show marked improvements, particularly PERL, where the robot consistently follows the desired path, emphasizing its real-time adaptability and precision. Overall, the PERL model outperforms the Fixed Physical and NN models in speed and position control, particularly in dynamic adjustments and precise tracking.

In Figs. 10-12, the first row illustrates the speed tracking performance of three robots under three different models. In the fixed physical model, the robot speed fluctuates significantly around the desired speed, demonstrating lower stability and control precision. The online NN model shows improvement with smaller fluctuations, though some deviations persist. The online PERL model excels, with the robot speed closely following the desired trajectory and minimal fluctuations, showcasing superior control accuracy and stability. The second row of subplots presents the corresponding position tracking data. In the fixed physical model, there are significant deviations between the actual and desired positions, especially in the later stages (time steps 250-300), highlighting clear position control errors. The online NN model, through real-time adjustments, improves position control across all three robots. The online PERL model, however, nearly perfectly tracks the desired position with minimal error, demonstrating its powerful capability in dynamic platoon control.

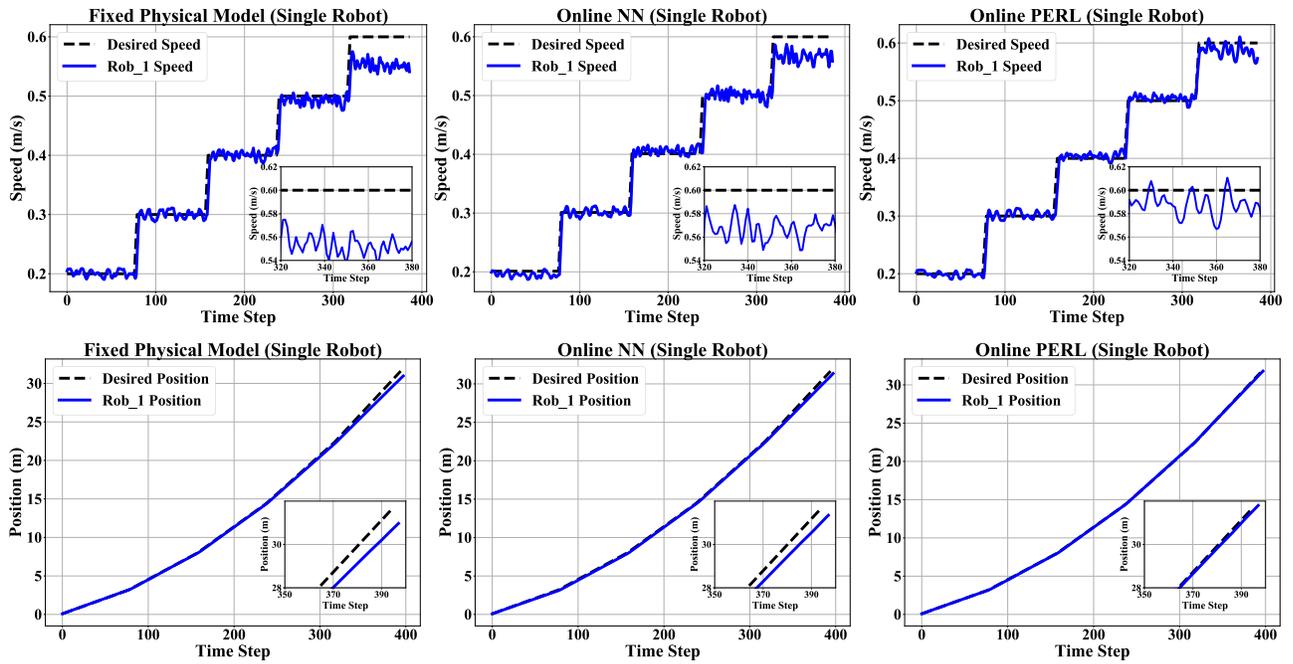

**Figure 9.** Speed and position comparison in single robort control.

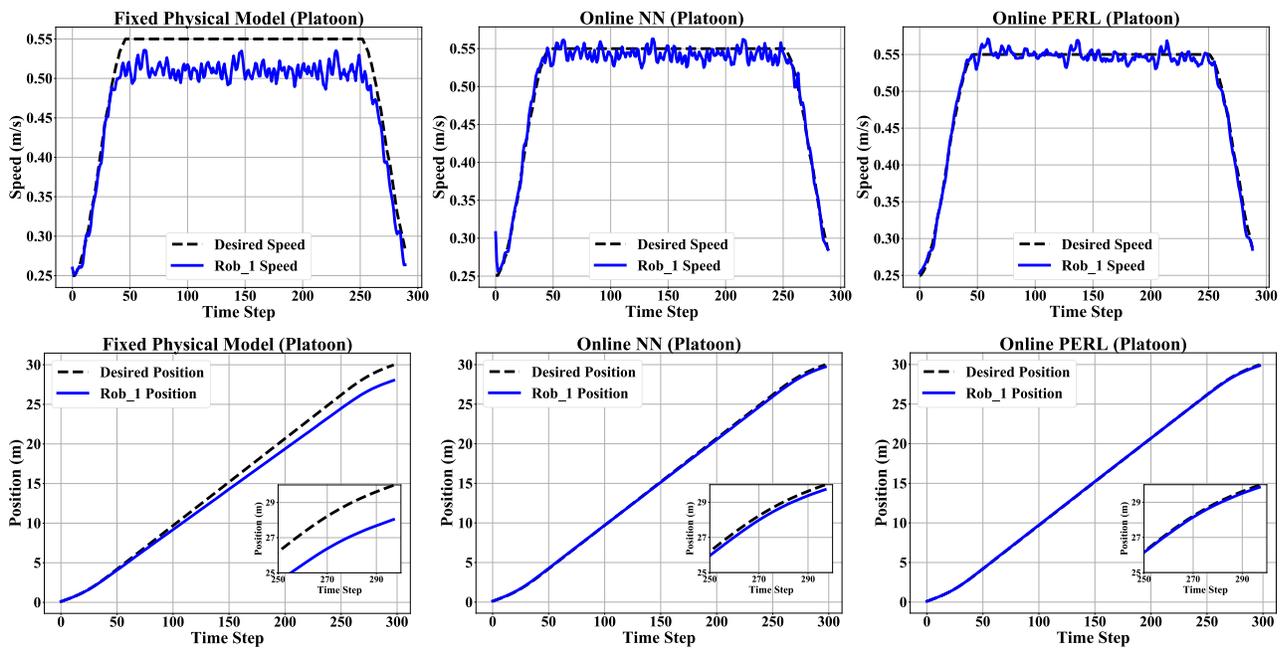

**Figure 10.** Speed and position comparison for robot 1 in platoon control.

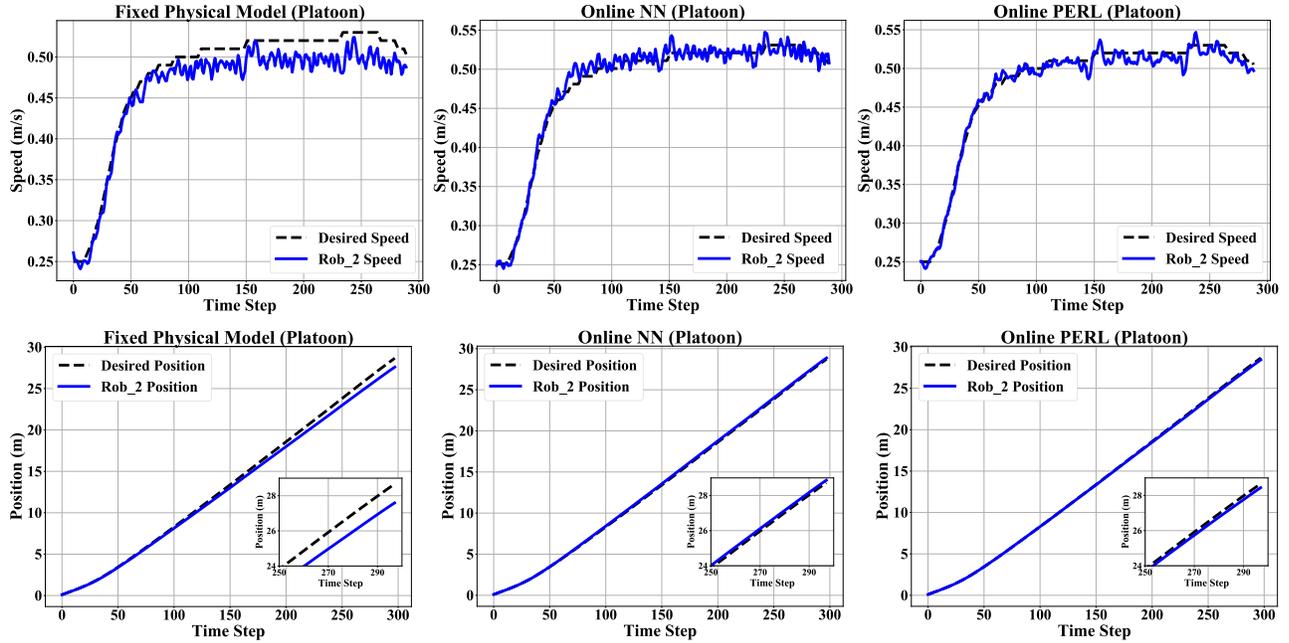

**Figure 11.** Speed and position comparison for robot 2 in platoon control.

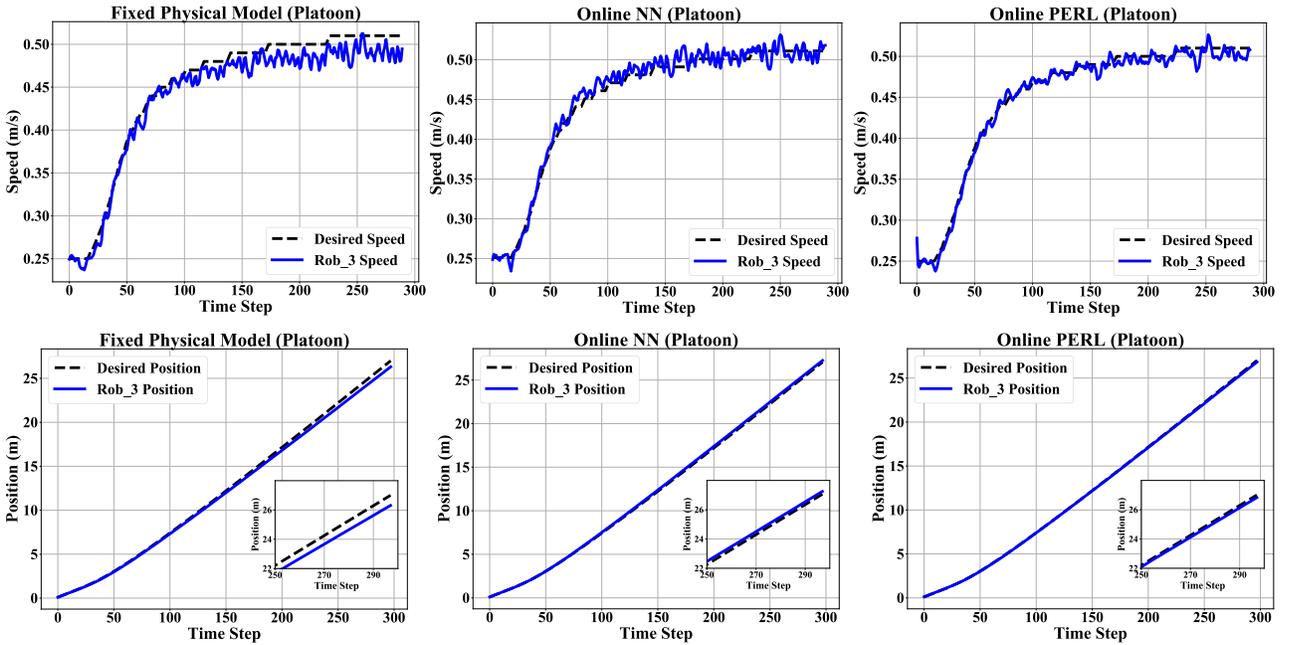

**Figure 12.** Speed and position comparison for robot 3 in platoon control.

The comparisons in **Figs. 10-12** indicate that the online PERL framework significantly outperforms both the fixed physical model and the online NN model in both speed and position control, showcasing exceptional performance in multi-robot platoons. The PERL model offers higher precision, stability, and responsiveness, proving its immense advantage in tasks requiring precise control.

Table 4 Control performance with different models in the field experiments.

| Test | Single Robot | | | Platoon (Robot 1, 2, 3) | | | | | | | | |
|---|---|---|---|---|---|---|---|---|---|---|---|---|
| | | | | Robot 1 | | | Robot 2 | | | Robot 3 | | |
| Metrics | Physic | NN | PERL | Physic | NN | PERL | Physic | NN | PERL | Physic | NN | PERL |
| S. MSE ($m^2/s^2$) | 0.0007 | 0.0005 | 0.0003 | 0.0014 | 0.0002 | 0.0002 | 0.0005 | 0.0013 | 0.0002 | 0.0003 | 0.0002 | 0.0002 |
| P. MSE ($m^2$) | 0.0796 | 0.0436 | 0.0017 | 1.2525 | 0.0202 | 0.0019 | 0.2818 | 0.0156 | 0.0090 | 0.1075 | 0.0211 | 0.0047 |

Table 4 summarizes the control performance of different models in single-robot and robot-platoon field tests, reporting mean squared error (MSE) values for speed (S.MSE) and position (P.MSE). The PERL model consistently achieved the lowest MSE values, demonstrating superior accuracy and performance. In single-robot tests, the PERL model recorded an S.MSE of 0.0003 and a P.MSE of 0.0019, reducing speed and position errors by 57.14% and 97.86% compared to the physical model, and by 40.00% and 96.10% compared to the NN model. This trend persisted in the platoon tests, where the PERL model outperformed its counterparts. For robots 1, 2, and 3, the PERL model also outperformed others, with average S.MSE and P.MSE values of 0.0002 and 0.0052 across three robots, reducing errors by 72.73% and 99.05% against the physical model, and by 64.71% and 72.58% against the NN model. These results highlight the exceptional performance of the PERL framework, offering lower control errors and higher precision compared to the fixed physical and NN models in both single robot and platoon control.

## 6. Conclusions

This paper introduces an online adaptive PERL controller that combines a physical model with residual learning, showcasing exceptional capability in addressing the challenges of dynamic and unpredictable CAV platoon systems. The PERL controller, using NN-learning to adjust the physical controller's output, minimizes transitional disturbances. It integrates high-precision residual predictions and adaptability, incorporating a physical model with inertial delay and using velocity as the control output for multi-objective optimization. Online residual learning effectively addresses disturbances from complex environments and vehicle dynamics. Simulation results reveal that trajectories generated by the online PERL controller have significantly lower errors compared to the physical model, with average cumulative absolute position and speed errors reduced by 58.5% and 40.1%, respectively, and by 58.4% and 47.7% compared to NN models. Tests on a reduced-scale robot car platform further confirm the superiority of the adaptive PERL controller, with position and speed errors reduced by 72.73% and 99.05% compared to the physical model and by 64.71% and 72.58% compared to NN models. PERL's online parameter updates in response to external disturbances lead to marked improvements in vehicle control, with the system demonstrating rapid convergence and high accuracy, ensuring platoon stability under various conditions. Future work will explore optimal physical control models and residual learning methods for CAV control and test the PERL on large-scale platforms to evaluate its effectiveness and robustness against real-world disturbances.

**CRediT authorship contribution statement**
**Peng Zhang:** Methodology, Software, Data processing, Writing original draft. **Heye Huang:** Methodology, Data curation, Writing. **Hang Zhou:** Visualization, Validation. **Haotian Shi:** Code review, Writing review. **Keke Long:** Conceptualization, revising original draft. **Xiaopeng Li:** Conceptualization, Supervision, Writing review & editing.

**Code availability**
Code availability is available at https://github.com/CATS-Lab/AV-Control-PERL_Platooning.

**Acknowledgment**
This work was sponsored by the Center for Connected and Automated Transportation (CCAT) project "Traffic Control based on CARMA platform for maximal traffic mobility and safety", and also supported by the U.S. National Science Foundation (NSF) under Grant No.2313578.

**Data availability**
Data will be made available on request.